\documentclass[11pt,a4paper]{article}
\usepackage[hyperref]{acl2020}
\usepackage{times}
\usepackage{latexsym}

\usepackage{comment}
\usepackage{graphicx}

\usepackage{microtype}

\aclfinalcopy 

\makeatletter
\newcommand{\printfnsymbol}[1]{%
  \textsuperscript{\@fnsymbol{#1}}%
}
\makeatother

\title{Data Driven Content Creation using Statistical and Natural Language Processing Techniques for Financial Domain}

\author{Ankush Chopra\thanks{*Equal Contribution}\\
  Fidelity Investments, AI CoE\\
  Bengaluru, India\\
  \texttt{ankush01729@gmail.com} \\\And
  Prateek Nagwanshi\printfnsymbol{1}\\
  Fidelity Investments, AI CoE\\
  Bengaluru, India\\
  \texttt{kunal.pn@gmail.com} \\\And
  Sohom Ghosh\printfnsymbol{1}\\
  Fidelity Investments, AI CoE\\
  Bengaluru, India\\
  \texttt{sohom1ghosh@gmail.com}\\
}

\date{}

\begin{document}
\maketitle
\begin{abstract}
Over the years customers’ expectation of getting information instantaneously has given rise to the increased usage of channels like virtual assistants. Typically, customers try to get their questions answered by low-touch channels like search and virtual assistant first, before getting in touch with a live chat agent or the phone representative. Higher usage of these low-touch systems is a win-win for both customers and the organization since it enables organizations to attain a low cost of service while customers get served without delay. In this paper, we propose a two-part framework where the first part describes methods to combine the information from different interaction channels like call, search, and chat. We do this by summarizing (using a stacked Bi-LSTM network) the high-touch interaction channel data such as call and chat into short search-query like customer intents and then creating an organically grown intent taxonomy from interaction data (using Hierarchical Agglomerative Clustering). The second part of the framework focuses on extracting customer questions by analyzing interaction data sources. It calculates similarity scores using TF-IDF and BERT \cite{devlin2018bert}. It also maps these identified questions to the output of the first part of the framework using syntactic and semantic similarity.

\end{abstract}


\section{Introduction}
\label{sec:intro}
In the current age information is the key to everything. Faster access to correct information has become an essential need. Customers interact with service providers through various channels looking for information. Information that is being sought, help customers make the right decision or finish a task/transaction that they are intending to do.

Customers use channels including but not limited to on-site search, call, live chat, virtual assistant, and emails. Customers use these channels in an order where they go from simple to more complex channels successively if their information need is not fulfilled by simpler channels. Search is the simplest channel since it is always available, and customers can look for the information very quickly. Next, comes virtual assistant, since even this has high availability, but it may or may not have answers to all the questions. Next, come chat, call, and email. Information provided through a low complexity channel tends to result in not only faster customer query resolution but also proves to be cost-effective for the organization.

Customers interact differently while using these different channels. The nature of the queries from these different channels also differs a lot due to the idiosyncrasies of the channels. For example, the search will have noticeably short inquiries that would not be a well-formed customer question. Virtual assistants get proper context-independent questions. Live chat and call tend to have pleasantries and general chit-chat along with the inquiry that may be broken into multiple sentences.

Often information that is being sought also has a sense of recency attached to it, and a large part of the customer base might also have similar queries. These queries are mostly related to a product or service that an organization offers, a recent event like new tax rules or initial public offering (IPO), specific to customers’ current state and so on.

In this paper, we propose a framework to identify the hot topics/intents and questions related to these hot topics that customers are seeking answers to. The framework is divided into 2 parts, first where we identify the hot topics/intents and iteratively cluster those to create an organic intent hierarchy. The second part focuses on identifying and extracting the customer questions from the interaction data and mapping those to the topics or topic clusters. This also helps content writers to get an idea regarding the topics on which they need to write articles for satisfying the customers' curiosity.


\textbf{Our contributions}:  We have developed two NLP based models i) First one performs hierarchical clustering of customer interactions iteratively ii) Second one maps questions to the cluster heads with descending order of frequency.

The next section will consist of a narration of prior works relating to this. After that, we will describe the problem, which will be followed by a section describing the solution methodology. We will shed some light on experimentation and results after the methodology section. Finally, we will conclude the paper by talking about the future enhancements we are about to bring to the framework.

\section{Related Works}
\label{sec:related works}
Tsai et al. in their paper \citep{tsai:ecir2013} described an approach that mined information from financial reports of a set of organizations. Using this information, it ranked these organizations as per their risk levels. They used learning-to-rank algorithms and benchmarked their model with a regression-based one. They also used Term Frequency Inverse Document Frequency (TF-IDF) \citep{tfidf} matrix of unigrams as features and trained a Support Vector Machine (SVM) \citep{svmvapnik1995support} model. In the paper \citep{criag2019}, Lewis et al. discussed how different Natural Language Processing based approaches (like Word Counts, Latent Dirichlet Allocation etc.) can be used to analyze Financial Texts.

A process of predicting Financial Markets (in terms of volatility, returns and traded volumes) using Google’s search queries from four English speaking countries has been narrated by Perlin et al. in their paper \citep{perlin:2017}. They studied how search patterns of some specific finance related terms on Google were related to the behaviour of Financial Markets. They used vector autoregression for modelling. Similar work had been presented by Mao et. al in their paper \citep{mao2011predicting}. They analyzed how various sentiment scores obtained from Tweets, Google search volume, Negative News Sentiment were related to the market conditions in terms of trading volumes, gold prices and so on.  

Li et al. in their paper \citep{li2014deep} proposed a solution for personalizing web search results using semantic and click-based features. They used three types of models namely XCode, Deep Structured Semantic Model and Convolutional Deep Structure Semantic Model. This led to a better ranking of the search results. However, this solution is generic and does not specifically deal with the financial domain. 

Litvak et al. in their paper \citep{litvak-etal-2020-hierarchical} described an approach of creating hierarchical summaries from financial reports. They used CODRA framework \citep{joty-etal-2015-codra} for discourse parsing.

It is interesting to note that all the papers described above (except the last one) dealt with predicting events/conditions of Financial Markets. None of these works relates to how customer interactions related to finance through different channels like search, calls, live chat, chat with virtual agents can be used to identify what their needs are and help the content writers prioritize their work. 

\section{Problem Statement}
\label{sec:problem}
Given a set, I = \{i\textsubscript{1}, i\textsubscript{2} … i\textsubscript{n}\} consisting of user interactions (i\textsubscript{1}, i\textsubscript{2} … i\textsubscript{n}) from the financial domain, we develop a system capable of extracting trending topics/intents and questions from it.

\section{Data and Preprocessing}
\label{sec:data}
The dataset consists of interactions that users had with Fidelity Investments\footnote{fidelity.com} through various channels like search, live chat and call over a period of one year (June 2020 - May 2021). The data distribution is mentioned in Table \ref{tab:data}. We list our data collection and pre-processing steps here.
\begin{table}
\centering
\begin{tabular}{|l|l|l|}
\hline
\textbf{Chanel} &
  \textbf{\begin{tabular}[c]{@{}l@{}}\# Interactions\end{tabular}} &
  \textbf{\begin{tabular}[c]{@{}l@{}}\# Distinct \\Interactions\end{tabular}} \\ \hline
Search        & 29.9 M & 2.9 M  \\ \hline
Live Chat     & 12.2 M & 96.7 K \\ \hline
Call          & 31.4 M & 31.4 M      \\ \hline
\end{tabular}
\caption{Data distribution}
\label{tab:data}
\end{table}
\subsection{Calls and Chats data}
Customer care representatives (reps) record a summary note for a fraction of customer calls that come in. We call these summaries - 'repnotes'. Due to their idiosyncrasies, reps introduce a lot of variations into the repnotes even if they mean the same thing. For example, ``\textit{customer contacted to reset the password}" and ``\textit{customer asked for the help with account reset}" are worded differently even if the meaning is the same. We used customer call data for training the intent models. We had both repnotes and transcripts for a portion of these calls. This made the call data ideal for modelling, as it gave almost ready to use training data. Call transcripts were input into the model and repnote was used as the summary. We applied certain filter conditions to select the right modelling data:
    \begin{itemize}
        \item We took data only from inbound calls since it’ll be applied to inbound interactions later.
        \item Since we were interested in extracting short intents of the calls, we only took calls where repnotes of length 6 or less were present.
    \end{itemize}

We performed cleaning steps on the call transcripts and repnotes mentioned below: 
\begin{itemize}
    \item Most call transcripts contained system messages (like \textit{“party has left the session”} and \textit{“This conversation is get recorded”}). They carried certain system meta-information only and were removed.
    \item Call transcript and repnotes were converted  to lower case
    \item Personal information of the customers which were masked were removed from the call transcripts and repnotes.
    E.g. \textit{[name], [number redacted], [unk]}, etc.
    \item Non-vocalized noise transcription which had been performed on call transcripts was removed
    \item Contraction replacement was performed on both call transcript and repnotes to normalize the text like ``I’ve" and ``We haven't" were expanded to ``I have" and ``We have not".  
    \item Non-informative prefixes like ``customer contacted" or ``customer asked" were removed from repnotes since they did not add any value. 
\end{itemize}
 Chat data comprised interaction between customers and representatives through text messages. We perform similar cleaning on the chat data as well.
 
\subsection{Search Data}
Search logs needed to be filtered and preprocessed before it is combined with the generated themes from call and chat. Since this data is from financial/ brokerage firm Fidelity Investments, the search log contained a lot of stock tickers, mutual fund tickers, names of the listed entities and so on. As these inquiries were used for getting the quote for respective stocks or funds, they were not relevant for the intent extraction and content creation process.  

Analysis of the search logs revealed that a small percentage of search queries were responsible for the majority of the searches. Search logs had an extremely long tail. We selected queries that were searched for at least 5 times. These queries constitute 97\% of the search volume. Furthermore, this made the search data manageable. 
\section{Methodology}
\begin{figure*}
\centering
  \includegraphics[width=14cm,height=6cm]{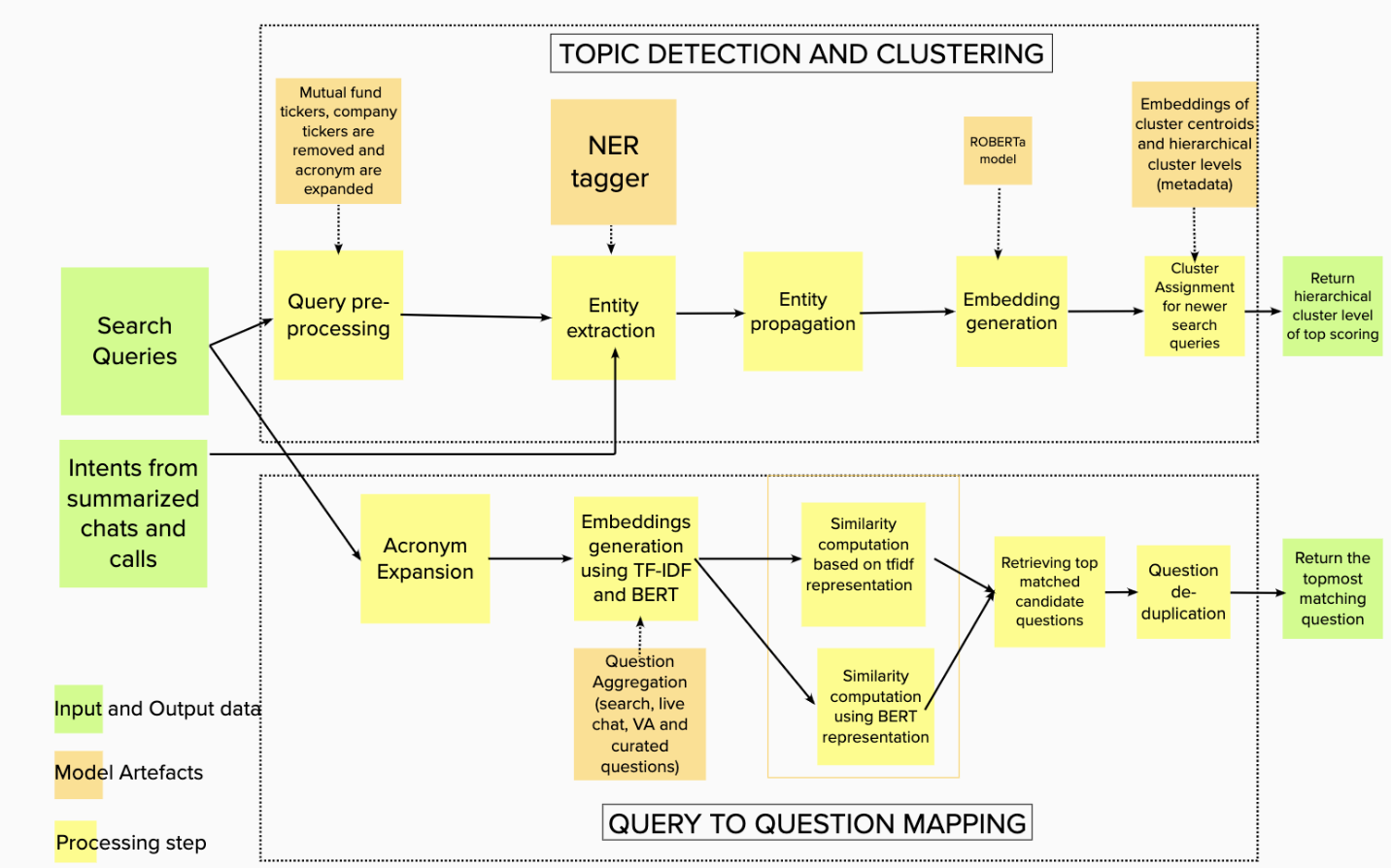}
  \caption{Flowchart representing our proposed approach}
   \label{fig:flow}
\end{figure*}

This work is divided into two major parts - identification of hot topics by creating  an intent tree organically and extracting popular questions from customer interactions. We initiate by describing the first of the two major parts of the framework. The intent tree created represents an organic hierarchy of topics/intents that customers were interested in. We used search queries, call transcripts and live chat from the customers to build this view. By nature, search queries were small and succinctly defined the intent of the customers in the majority of cases. Call transcripts and chat were verbose and had customer intent hidden somewhere in the body of the conversation. To be able to use all these communication channels in tandem, they should have had the same structure. Hence, we first worked towards extracting the customer intent from the calls and chats.

\subsection{Extracting Intents from Calls and Chats}
 We performed abstractive summarization of the calls, given that the nature of the relationship between rep-notes and call transcripts was inherently abstractive. The proposed model is based on 2 stacked Bi-LSTM (Long Short Term Memory) \citep{lstm} Sequence  to Sequence with Attention \citep{vaswani2017attention} architecture. We did not use any pre-trained embedding due to the nature of the task. Instead, we are learning the embedding while training for this task. 

This model was trained for 27 epochs and training terminated due to early stopping criteria. The model had categorical cross-entropy accuracy of 79\% on the validation set. This model is used to generate the customer intents behind all incoming calls and chat transcripts irrespective of whether they have repnotes or not. These generated repnotes look similar to short search queries and explain the intents behind the calls.

\subsection{Combining and clustering search and other interaction data sources}
Search queries were mostly specific and short in length. Thus, we used them as it is. Once data from all the channels were in a similar format, we proceeded to the pre-process and normalize them. We replaced acronyms with their definitions, removed repetitive words and any customer specific information. We then converted these intent phrases to embedding using the RoBERTa \citep{liu2019roberta} pre-trained model (768 dimensions). We made use of the sentence transformer \citep{sbert} to come up with sentence embeddings. We performed standardization and reduced the dimensions to 300 using Principal Component Analysis (PCA). This was done to make the data manageable.  

We used a Financial Named Entity Recognition (NER) module developed in-house to identify the business entities (product and services) present within customer intents. We divided the intents into smaller groups based on the entities that are present in them. This helped us to achieve cleaner clusters at the product and service level.  

We performed iterative clustering within each of the intent cohorts using agglomerative hierarchical clustering (4 levels). We first group the generated intents which had at least $x$ similarity and calculated cluster centroid for all the newly formed clusters. These cluster centroids were again clustered in the groups which had at least $x$ = $x-$ $\Delta x$ similarity. Depending on the business requirement we repeated the above steps multiple times. For example, if we needed 3 level cluster and the similarity threshold for first level clustering was x=0.85
and for subsequent levels, the criteria were relaxed by
 $\Delta x$=0.05. Then we had clusters with 0.85, 0.80 and 0.75 similarity thresholds. We used cosine similarity as the measure of affinity.

\subsection{Query to Question Mapping} 
Here, we propose an algorithm for mapping a search query to its probable questions present in an existing database. The question database contained all the questions that were being asked across different channels (live chat, virtual agent, and search). In addition to it, a set of curated questions were also present.  Showing questions for a given search query gave an overall picture of what questions were being asked related to the keyword which was searched. 

There were several steps involved in mapping a query to its candidate questions.  The following lines explain each of these steps.\\
\textbf{Question Detection:} We used search queries, live chat, and virtual agent messages for detecting questions using a question detector algorithm developed in-house\\
\textbf{Question Aggregation:} Due to the different nature of the channels, there were slight variations in detected questions. We needed some bit of pre-processing and normalization of these questions to get the overall volume of certain types of questions. In addition to simple pre-processing and we also applied acronym expansion (like IPO to Initial Public Offering), to get a unified set of questions along with their aggregated count.\\
\textbf{Embedding Creation:} For matching a query to its probable questions, we needed to find the similarity between the search query and questions. We further calculated the distance between the query embedding matrix and the embedding matrix created from questions. The embedding techniques used in this work constitute TF-IDF and Sentence BERT. TF-IDF had been used for keyword-based embedding and Sentence-BERT (Bidirectional Encoder Representations from Transformers) \citep{sbert} had been used for creating semantic embedding. \\
\textbf{TF-IDF:} We performed different pre-processing steps to clean the data such as acronyms were replaced by their definitions and customer-specific information was removed and so on. All questions had been lemmatized to their base word. Moreover, after removing all stop words these normalized questions were used for creating the TF-IDF (uni-grams and bi-grams were used) matrix which was then used for matching the vectors created from questions.\\
\textbf{SentenceBERT:} \citep{sbert} has presented a modification of BERT \citep{devlin2018bert} model that uses Siamese network to derive semantic sentence embeddings that can be used to calculate cosine similarity between sentences. For our experiments, we had used Sentence Transformer model based on BERT and RoBERTa.
We performed all the pre-processing steps which have been mentioned in the previous section to clean the data excluding the lemmatization step. After removing stop words, the normalized question was used for creating the embedding of 768 dimensions from Sentence-BERT. It was used for finding the similarity with embedding created for the query using Sentence-BERT.\\
\textbf{Mapping Query To Questions:} For finding probable questions for a given search query, we calculated the cosine similarity between question embeddings (created using TF-IDF and Sentence -BERT) and query embedding. All questions beyond the similarity threshold (0.89 for TF-IDF and 0.86 for BERT) were collected. There was only a slight variation among all such questions. To detect near-duplicate questions, clustering was done and question with high frequencies was assigned as cluster head. For a given search query only the cluster heads were displayed. 
We present the entire workflow in the Figure \ref{fig:flow}.

\section{Experimentation and Results}
We conducted separate experiments for each of the modules as narrated in the following parts.

\subsection{Extraction of Intents from Interaction Data}
Firstly, we started with the pre-processing and cleaning of the call transcripts and repnotes. We removed system noise, transcription-induced noise, and masked tokens from both sources. We also performed case normalization, contraction replacement on both data sources. We then tokenized both cleaned transcripts and repnotes separately. We divided the data into the train, validation, and out-of-time test set. Close to 35\% of the cleaned repnotes were of length 6 and less, close to 40\% were of length 7 to 17 and remaining were of more than 17 words. 90\% of the call transcripts were of length 450 words or less. For modelling, we considered all the calls with less than or equal to 450 tokes in the transcripts where repnotes of length 6 or less were present.  

\begin{figure}[ht]
\includegraphics[width=5cm]{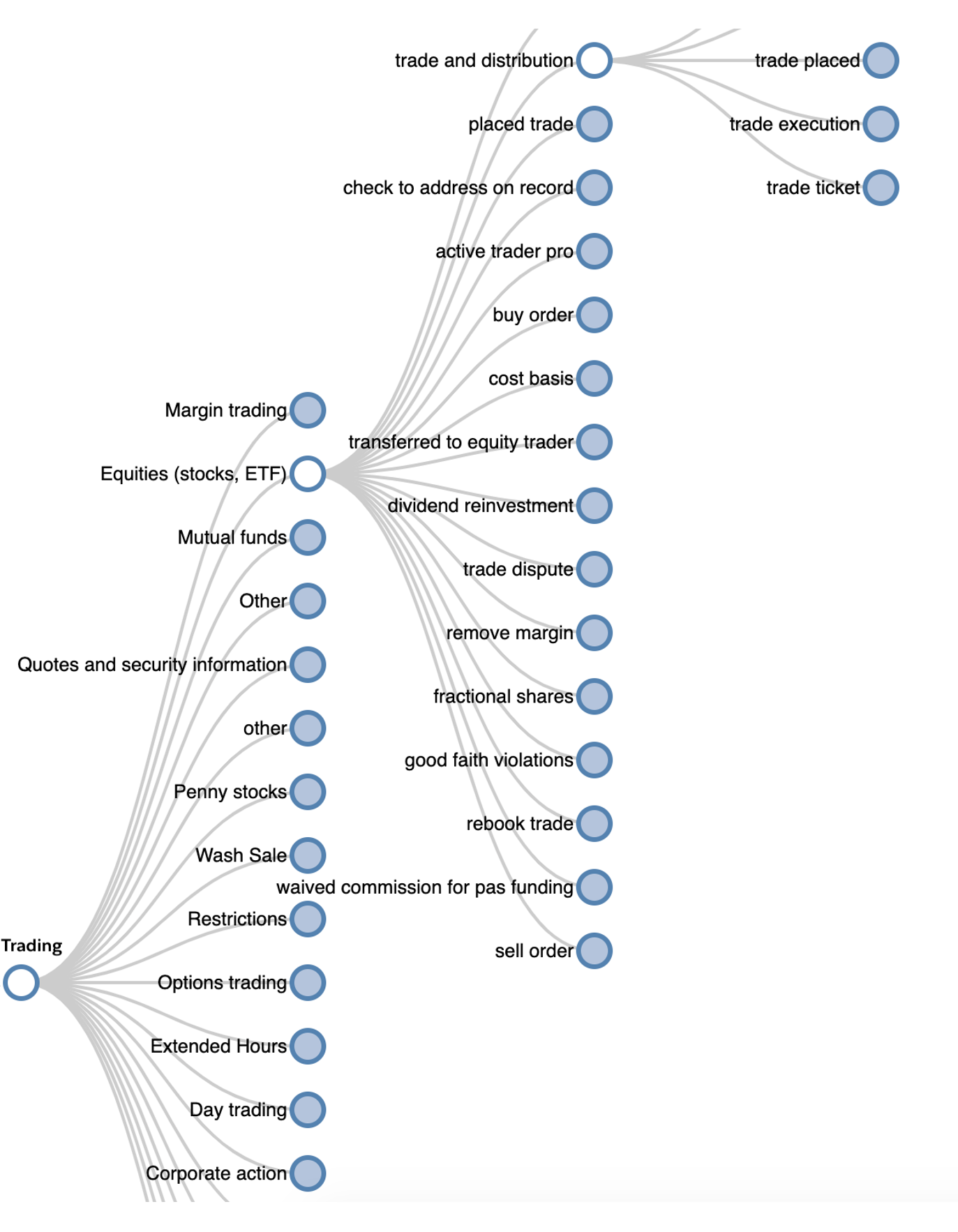}
\caption{Hierarchical Tree Visualization}
\label{fig:tree}
\end{figure}

 
 Since this is an abstractive summarization problem, we chose to use sequence-to-sequence (s2s) Recurrent Neural Networks (RNN) architecture for modelling this. We chose LSTM \citep{lstm} variant of RNNs, since it is proven effective in capturing longer term dependency. We started off with s2s architecture without attention. It did not yield good results. Upon closer examination, it was clear that performance was getting worse with an increase in the input sequence length.\\
Next, we tried the s2s with attention \citep{vaswani2017attention} since it is known to work better for longer input sequences. This significantly helped in improving the model’s performance. We experimented with the common LSTM hyperparameters like the number of layers in LSTM, gradient clipping, dropout, recurrent dropout etc. We also tried pre-trained transformer-based fine-tuning using BERT andT5 but they did not perform as good as LSTM for the validation set. It is probably due to the poorly transcribed data with improper sentence and grammar structure. Inference using transformer-based models was slow as well. We used s2s with attention model for generating the interaction intent after comparing the performance of the above model using manual validation.\\
We combined the generated intents with the customer search queries. In our context, interactions belong to two broad categories, customers looking for a quote of a traded entity like a company or mutual fund and seeking information on any product, service, or recent transaction. We had taken the intents belonging to the latter case since the former did not require new content to be written for it to be answered. We also normalized the product name variations, acronyms, contractions, and removed the repetitive words and phrases.\\
Even after normalization and cleaning, the number of intents was in the range of thousands. We decided to iteratively group these intents into homogeneous clusters. We began with clustering the cleaned intents into groups where constituent members were highly similar to each other. The intent with the highest interaction count was chosen as cluster name. We also calculated the centroid for all the newly formed clusters and clustered them again. This time we reduced the similarity threshold marginally. This gave us a new cluster of clusters. We similarly named these clusters as before. By doing this iteratively a few times, we ended up with few hundred clusters that had links to one of the products or services offered by the organization. We used agglomerative hierarchical clustering since the number of clusters were not known apriori.\\
We tried few different featurization techniques to get the best clusters. Firstly, we created features using TF-IDF for the cleaned intents and tried clustering these into smaller buckets. We then used PCA to make the embedding size manageable (300 dimensions were retained). As expected, these clusters were failing to capture the semantic similarity. Intents like “account reset” and “password change” were not getting grouped.\\
Next, we generated the vector representation of the intent phrases using Word2Vec \citep{word2vec}. We tokenized the phrases into words and took the average of the word embeddings present in the intent phrase. It performed better than TF-IDF based approach, but it wasn’t doing well where a word had a different meaning in a different context. For example, “option” in “payment option” and “option trade” has a completely different meaning which word2vec based model was not able to disambiguate. Additionally, it was not helping with misspellings as it provides word-level embeddings.\\
Further, we experimented with Transformer based embedding models like BERT \citep{devlin2018bert} and RoBERTa \citep{liu2019roberta}. We used Sentence Transformers to get the phrase embedding for the intent phrases while using BERT or RoBERTa in the back-end. We could see that RoBERTa performed better than BERT. RoBERTa (large base model) performed the best when we looked at the silhouette scores to compare the performance of different clustering methods. Manual validation also indicated that transformers based embedding models were able to capture the semantic and syntactic similarity better than the other models. Table \ref{tab:silhouttescores} shows the model performance. The organically generated hierarchy of intents has been depicted in Figure \ref{fig:tree}.

\begin{table}
\centering
\begin{tabular}{|l|l|l|}
\hline
\textbf{cluster level} & \textbf{search query} & \textbf{calls} \\ \hline
\textbf{0}             & 0.09                  & 0.48           \\ \hline
\textbf{1}             & 0.10                  & 0.35           \\ \hline
\textbf{2}             & 0.08                  & 0.23           \\ \hline
\textbf{3}             & 0.07                  & 0.12           \\ \hline
\end{tabular}
\caption{RoBERTa based Silhoutte Scores}
\label{tab:silhouttescores}
\end{table}

\subsection{Extraction of Questions from Interaction Data}
Query to question mapping is a task to assign 'n' candidate questions to a search query. The value of 'n' differs for each query depending upon the questions being asked around that search query and the similarity between search query and questions. We performed our experiments using different kinds of embeddings such as TF-IDF, BERT and combined output of TF-IDF and BERT based similarity. We started with normalizing all the questions, which included pre-processing and expansion of acronyms with their definition.  Since we did not have tagged data to evaluate this model, we manually evaluated it. We chose the top 10,000 search queries and created the probable questions mapping for them.  We used a sample of 1000 records for manual evaluation.\\
Initially, we started with TF-IDF for creating features of aggregated questions. We performed case normalization and removed noisy characters like `?',`.' from the end of questions. Then we replaced the acronyms with their definition and after that, we converted the words with their lemmatized version. We further tokenized the questions and removed the English stop words like `a', `the', etc. Using the normalized version of questions, we had created a TF-IDF matrix (TF-IDF\textsubscript{ques}). For search queries, we also performed the same pre-processing and created the TF-IDF matrix (TF-IDF\textsubscript{query}) for search queries as well.  We calculated the cosine similarity between the question TF-IDF\textsubscript{ques} and TF-IDF\textsubscript{query} and evaluated the performance at different levels of threshold.  This model was giving descent performance on queries that had some matching words with the question. However, queries like `withdrawal' did not have any matching questions even if they contained questions related to `withdrawal'. So, this TF-IDF model was unable to capture the semantic aspect.
For covering the semantic aspect as well, we further experimented with sentence transformers to get the phrase level embedding using BERT. We used the normalized version of the question for creating the embedding matrix SBERT\textsubscript{ques}  and using the normalized query created SBERT\textsubscript{query}. Finally, we calculated the cosine similarity between SBERT\textsubscript{query} and SBERT\textsubscript{ques} to get the questions for a query at the various levels of threshold. We also tried changing the BERT model with RoBERTa for creating SRoBERTa\textsubscript{ques} for questions and SRoBERTa\textsubscript{query} for queries. We calculated the cosine similarity between SRoBERTa\textsubscript{query} and SRoBERTa\textsubscript{ques} to get the set of candidate questions for a query. The model was able to capture most of the variations of questions present in the database. Still, the coverage was very low compared to the TF-IDF method. Finally, we combined the output of TF-IDF and Sentence Transformers with RoBERTa \cite{liu2019roberta} at the back-end to get the final candidate set of questions for a given query. It captured the variations from semantic embedding and word-based matching using TF-IDF. Since we did not have the tagged data to evaluate this task. We chose to perform a manual evaluation and calculate the precision score for evaluating the model. We present the individual performance of both the methods at various thresholds in Table \ref{tab:query-to-ques}. We calculated the precision score based on a manual evaluation of 1000 query question pairs. After looking at the precision score we concluded that both the methods were performing well for threshold $>$= 0.8. After doing a failure case analysis, we saw that in plenty of cases where TF-IDF was failing, and Sentence-BERT was performing good and vice-versa. 
We tried combining TF-IDF that capture the bag-of-words aspect and Sentence-BERT that covers the variations and semantic aspect of the language. We saw if we combine both the models then we are getting much broader coverage with plenty of variations of questions for a query, which is not the case with individual semantic or word-based methods. Some examples of search queries and corresponding questions extracted are presented in Table \ref{tab:query-to-ques}. Finally, we clubbed the output of TF-IDF and Sentence Transformers (with RoBERTa at the back-end) to get the final candidate set of questions for a query.

\begin{table}
\centering
\begin{tabular}{|l|l|l|}
\hline
\textbf{Th} & \textbf{Pr TFIDF (\%)} & \textbf{Pr SBERT(\%)} \\ \hline
0.4         & 84.11            & 83.80            \\ \hline
0.5         & 87.00            & 84.20            \\ \hline
0.6         & 87.50            & 87.70            \\ \hline
0.7         & 87.56            & 87.32            \\ \hline
0.8         & 90.70            & 88.93            \\ \hline
0.9         & 92.39            & 92.30            \\ \hline
\end{tabular}
\caption{Comparison of Precision(Pr) Scores of Query to Question Model trained using TF-IDF and Sentence BERT based similarity for different thresholds (Th)}
\label{tab:resultsques}
\end{table}

\begin{table}
\centering
\begin{tabular}{|l|l|}
\hline
\textbf{SQ} & \textbf{EQ}    \\ \hline
tax form                & where do i view my tax form?    \\ \hline
tax form                & when do tax forms get sent out? \\ \hline
tax form                & can i get my tax form?          \\ \hline
direct deposit          & how to edit direct deposit?     \\ \hline
direct   deposit        & what is a direct deposit?       \\ \hline
\end{tabular}
\caption{Search Queries (SQ) and Extracted Questions(EQ)}
\label{tab:query-to-ques}
\end{table}

\section{Conclusion and Future Works}
In this paper, we discussed the process to extract insights from customer interactions by clustering them hierarchically. It led to the creation of an intent hierarchy organically. We further narrated a methodology to mine queries from these interactions and rank them. Various regulations do not permit machine generated answers to be directly given to the customers. This is specifically the case in sectors like finance and healthcare. Hence decoupling of the answer generation from the question mapping was needed to comply with the regulation. Our proposed system would help content writers efficiently identify the topics and questions which are being asked by a large number of customers. Once they write answers to these questions, a system like the one described in \citep{mlnlpchopra} could be used to serve these answers directly to the users through channels like search, Virtual Agents (Chatbots) and so on.

In future, we would like to incorporate market events to decide the prioritization of content making process. Furthermore, we would like to remove queries that already have enough content. We want to assign more priority to those topics which the customers searched for but did not lead to any fruitful results. Lastly, we want to do an extensive evaluation using external parameters like measuring the number of searches that are not followed by a call, reduction in call volumes and so on.





\bibliography{acl2020}
\bibliographystyle{acl_natbib}








\end{document}